\documentclass[letterpaper]{article} 
\usepackage{aaai2026}
\usepackage{times}  
\usepackage{helvet}  
\usepackage{courier}  
\usepackage[hyphens]{url}  
\usepackage{graphicx} 
\usepackage{natbib}  
\usepackage{caption} 
\usepackage{algorithm}
\usepackage{algorithmic}
\usepackage{pifont}

\usepackage{booktabs}
\usepackage{amsmath}
\usepackage{subcaption}
\usepackage{array}
\usepackage{multirow}
\usepackage{makecell}
\usepackage{ragged2e}
\usepackage{longtable}
\usepackage{xcolor}
\usepackage{comment}

\usepackage{newfloat}
\usepackage{listings}
\DeclareCaptionStyle{ruled}{labelfont=normalfont,labelsep=colon,strut=off} 
\floatstyle{ruled}
\newfloat{listing}{tb}{lst}{}
\floatname{listing}{Listing}
\title{Improving Implicit Discourse Relation Recognition with Natural Language Explanations from LLMs}
\author {
    Heng Wang\textsuperscript{\rm 1},
    Changxing Wu\textsuperscript{\rm 1}\thanks{Corresponding author.}\\
}
\affiliations {
    \textsuperscript{\rm 1} School of Information and Software Engineering, East China Jiaotong University, Nanchang, China \\
    2023218083500007@ecjtu.edu.cn, wuchangxing@ecjtu.edu.cn
}

\usepackage{bibentry}

\begin{document}

\maketitle

\begin{abstract}

Implicit Discourse Relation Recognition (IDRR) remains a challenging task due to the requirement for deep semantic understanding in the absence of explicit discourse markers.
A further limitation is that existing methods only predict relations without providing any supporting explanations.
Recent advances in large language models (LLMs) have shown strong reasoning capabilities in both deep language understanding and natural language explanation generation.
In this work, we propose a simple yet effective approach to distill the reasoning capabilities of LLMs into lightweight IDRR models to improve both performance and interpretability.
Specifically, we first prompt an LLM to generate explanations for each training instance conditioned on its gold label.
Then, we introduce a novel classification-generation framework that jointly performs relation prediction and explanation generation, and train it with the additional supervision of LLM-generated explanations.
Our framework is plug-and-play, enabling easy integration with most existing IDRR models. 
Experimental results on PDTB demonstrate that our approach significantly improves IDRR performance, 
while human evaluation further confirms that the generated explanations enhance model interpretability.
Furthermore, we validate the generality of our approach on sentiment classification and natural language inference.

\end{abstract}


\begin{links}
    \link{Code, Appendix}{https://github.com/nlper-hub/EIDRR}
\end{links}

\section{Introduction}


To fully understand natural language text, 
it is essential not only to comprehend the meaning of individual sentences but also to grasp the semantic relationships that link them, known as discourse relations (e.g., \emph{Comparison}).
Discourse relation recognition has attracted considerable attention in natural language processing (NLP) due to its potential to enhance both language understanding and generation 
\citep{li_composing_2020,tang_discourse_2021,hu_exploring_2023,li2024dialogue}.
Among these, implicit discourse relations—which lack explicit connectives such as \emph{but}, \emph{and}, or \emph{because}—pose a greater challenge for both humans and machines to identify.
Despite recent progress in NLP and the emergence of LLMs, implicit discourse relation recognition (IDRR) remains a challenging task \citep{chan_exploring_2024,yung_prompting_2024}.

\begin{figure}[t]
\centering
\includegraphics[width=8.5cm]{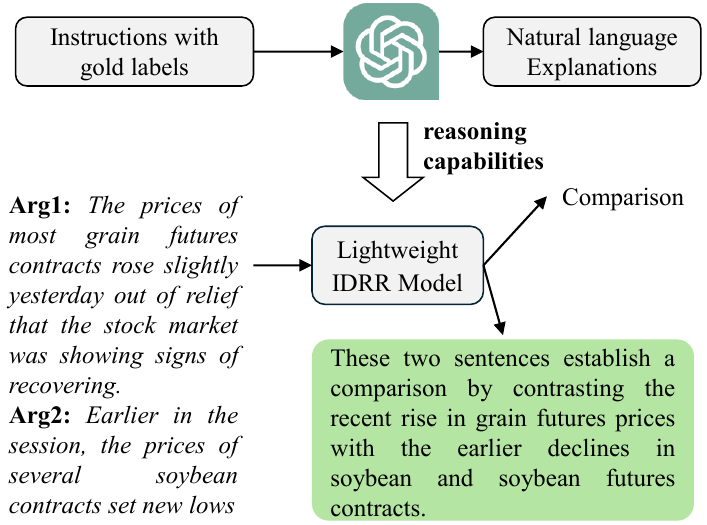}
\caption{
Our lightweight IDRR model distills LLM reasoning to predict discourse relations and generate interpretable explanations.
\textbf{Arg1} and \textbf{Arg2} are two arguments.
\label{fig:intro}}
\end{figure}

Implicit discourse relation recognition is typically formulated as a classification task, where the goal is to predict the discourse relation connecting two given arguments (e.g., sentences or clauses). 
With the advancement of deep learning, 
IDRR models that once relied on human-designed features \citep{park_improving_2012,rutherford_discovering_2014} have transitioned to neural network-based models.
Early research efforts primarily focused on designing task-specific neural network structures to better capture the semantics of arguments and model their interactions \citep{zhang_shallow_2015,bai-zhao-2018-deep,liu_importance_2020}.
Recently, researchers have enhanced IDRR models by leveraging relation hierarchies, 
adopting either generation models \citep{wu_label_2022} or contrastive learning approaches 
\citep{long_facilitating_2022}.
More recently, 
prompt-based IDRR models have demonstrated superior performance by utilizing pre-trained language models to predict discourse connectives and then mapping them to corresponding relations \citep{xiang_connprompt_2022,zhao_infusing_2023,zeng_global_2024}.
Despite their notable performance gains, current IDRR models output only predicted relations without providing any supporting explanations, which limits their interpretability and reduces user trust.

Recent advances in large language models (LLMs) have shown strong reasoning capabilities, 
including both deep language understanding and the ability to generate natural language explanations \citep{madsen2024self,bilal_llms_2025}.
However, empirical results indicate that LLMs, when used with prompting or in-context learning,
significantly underperform compared to lightweight IDRR models trained on human-annotated data \citep{chan_exploring_2024}, 
and also exhibit limited faithfulness \citep{miao2024discursive}.
While fine-tuning LLMs can improve performance, 
their heavy computational demands and high deployment cost limit their applicability in low-resource environments.
These limitations raise a natural question:
\emph{\textbf{can we transfer the reasoning capabilities of LLMs into lightweight models to enable more accurate and interpretable IDRR?}}

To address this question, 
we propose a simple yet effective distillation approach using natural language explanations as a bridge.
As shown in Figure \ref{fig:intro}, 
our approach consists of two main stages.
\textbf{First}, we prompt an LLM to generate natural language explanations for each training instance, 
conditioned on its gold label.
To obtain high-quality outputs, 
we draw inspiration from the Chain-of-Thought concept \citep{wei_chain--thought_2022}.
Specifically, we structure each explanation into two parts: the restatement of two arguments (serving as a thought step, not shown in Figure \ref{fig:intro} for clarity) and a rationale of their discourse relationship.
To some extent, these generated explanations not only offer a deep understanding of the input instances but also reflect the reasoning patterns inherent in LLMs.
\textbf{In the second stage}, we introduce a classification-generation framework equipped with a transformer module.
This framework learns to jointly predict discourse relations and generate explanations under the combined supervision of gold discourse labels and LLM-generated explanations.
To facilitate effective interaction between the two tasks, we adopt a multi-task learning strategy with a shared encoder.
The transformer module is introduced to alleviate the potential mismatch when using a shared encoder for both classification and generation objectives.
Our framework is plug-and-play and can be readily combined with most existing IDRR models by adding a simple transformer module and a generation component, with minimal changes required.

The primary contributions of our work are as follows:
\begin{itemize}
     \item We make an initial attempt toward interpretable IDRR by jointly predicting discourse relations and generating natural language explanations.
    
    \item We propose a plug-and-play classification-generation framework that facilitates the distillation of LLM reasoning capabilities into lightweight IDRR models.
  
    \item Experimental results and human evaluations confirm that our method significantly improves IDRR performance and delivers high-quality, interpretable explanations.
    
    \item We further validate the generalizability of our approach on two additional tasks: sentiment classification and natural language inference.
\end{itemize}

\section{Related Work}
\subsection{Neural IDRR Methods}
\label{subsec:idrrmodels}
Neural network-based methods have significantly advanced this area.
Previous researchers focused on designing diverse neural network architectures to effectively capture the interactions between arguments \citep{zhang2018learning,bai-zhao-2018-deep,zhang_context_2021}. 
During this phase, the utilization of pre-trained language models (PLMs) like BERT and RoBERTa significantly improved the performance \citep{liu_importance_2020}.
Meanwhile, some studies began exploring generation-based methods to better capture relational semantics.
For example, 
\citet{jiang2021not} proposed CG-T5, which treats IDRR as a generation task that jointly predicts discourse relations and generates sentences conveying their meanings.
Recently, researchers have enhanced IDRR models by integrating label hierarchy information. 
\citet{wu_label_2022} viewed multi-level IDRR as a label sequence generation task, enabling the effective utilization of inter-label dependencies.
\citet{long_facilitating_2022} and \citet{jiang_global_2023} introduced additional contrastive learning losses based on the hierarchical structure of labels.
More recently, prompt-based learning has emerged as a powerful paradigm, 
achieving the SOTA performance in IDRR.
\citet{xiang_connprompt_2022} and \citet{zhou_prompt-based_2022} created different prompt templates to encourage PLMs to predict connectives that link two arguments, which are then used to determine discourse relations.
\citet{zhao_infusing_2023}, \citet{chan_discoprompt_2023} and \citet{zeng_global_2024} infused hierarchical label information into prompt tuning.
The scarcity of annotated data in IDRR has driven the exploration of data-augmented approaches.
Researchers employed conditional variational autoencoders \citep{dou_cvae-based_2021} or large language models \citep{omura_empirical_2024} to create diverse, high-quality synthetic data to enhance the annotated dataset.
Others sought to leverage abundant explicit discourse data, naturally annotated with connectives, to pre-train prompt-based models \citep{wang_prompt-based_2023,liu_annotation-inspired_2023}.
Unlike existing methods that focus primarily on classification without providing explanations for their predictions, our approach enhances both predictive performance and model interpretability.


\subsection{Natural Language Explanations}

There is an increasing interest in providing natural language explanations for the decisions made by neural models \citep{lyu_towards_2024}. 
Based on manually-annotated explanations, 
\citet{liu_towards_2019} proposed a generative explanation framework for classification tasks, 
\citet{kumar_nile_2020} automatically generated explanations for each possible label of an instance and used them to make the decision with an explanation.
Recently, with explanations generated by LLMs, 
\citet{wang_reducing_2023} mitigated spurious correlations in aspect-based sentiment classification, 
\citet{ludan_explanation-based_2023} proposed an explanation-based fine-tuning method to enhance robustness, 
\citet{kroeger_context_2024} showed that their in-context explainer framework enables LLMs to generate explanations on par with the best explainers.
\citet{liu_enhancing_2025} and \citet{fan-etal-2025-improving} leveraged explanatory texts generated by LLMs to enhance the ability to handle complex semantic phenomena in multi-party dialogues, which improved performance in  discourse analysis.
Inspired by these studies, we present the first exploration of interpretable IDRR by leveraging explanations generated by LLMs.

\section{Method}
\subsection{Explanation-enhanced PDTB Construction}

PDTB \citep{prasad_penn_2008} is widely recognized as the largest and most commonly used corpus for IDRR. 
An implicit discourse instance in PDTB can be depicted as $(arg_1, arg_2, y)$, where $arg_1$ and $arg_2$ are two arguments and $y$ denotes the manually-annotated discourse label.
Recent studies have shown that LLMs like GPT3 \citep{brown_language_2020} can generate high-quality explanations for classification tasks, such as sentiment classification \citep{wang_reducing_2023,kroeger_context_2024}.
Motivated by this, we use LLMs to generate a natural language explanation $e$ for each PDTB instance, conditioned on its gold label.




\begin{figure}[t]
\centering
\includegraphics[width=8cm]{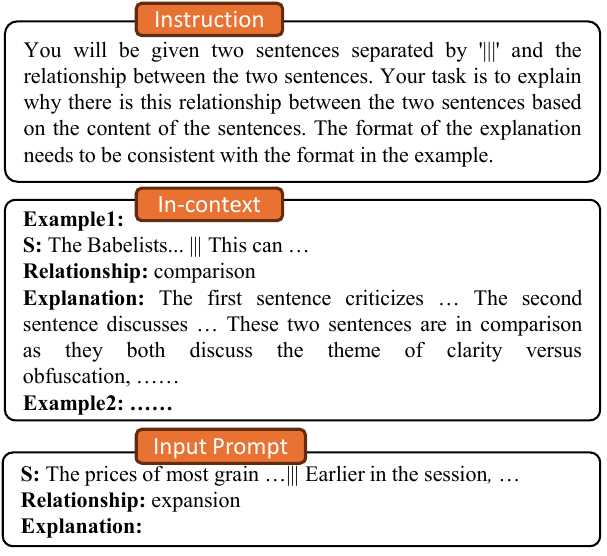}
\caption{The prompt template for explanation generation. 
An explanation consists of two parts: the restatement of two arguments, 
and a rationale of their discourse relationship.
\textbf{S} means two arguments.
\label{fig:template}}
\end{figure}   

In the explanation generation stage, we design a prompt template based on In-Context Learning \citep{brown_language_2020} to activate the understanding and inference capabilities of LLMs. 
As illustrated in Figure \ref{fig:template}, the template comprises three components: instructions, in-context examples, and the input prompt.
The instructions clearly explain the generation task and provide guidance on the output format.
In-context examples offer further support for LLMs to comprehend the task.
The input prompt embeds the arguments and the gold label of an implicit discourse instance for processing.
Following Chain-of-Thought, we format an explanation by first restating the two arguments (i.e., \emph{the first sentence criticizes..., the second sentence discusses}) 
and then explaining why they express a specific discourse relation (i.e., \emph{discuss the theme of clarity versus obfuscation...}).
Incorporating the restating of arguments serves two purposes:
1) it encourages the LLM to perform deeper semantic interpretation of the input arguments, and
2) it functions as a thought step in the Chain-of-Thought process, leading to improved explanation generation.


We compared the explanation quality of ChatGPT-4o, Qwen-long, and Ernie Bot-3.5-128K, and finally chose Qwen-long.
We randomly selected 1,000 instances and found that most explanations provide reasonable rationales for the given gold labels, with an average human evaluation score of 4.53 out of 5 (See evaluation criteria on \emph{P6}).
Therefore, we only manually corrected explanations of instances that were inconsistent with their labels.
We identified two main challenges in explanation generation.
1) Ethical concerns: LLMs reject instructions when encountering certain keywords. In such cases, we resort to alternative LLMs or manual processing.
2) Insufficient context: Short arguments often lead to incorrect or contradictory explanations. To mitigate this, we supplement the input with additional context from the source document.
In the end, we obtained explanations for 15,004 PDTB instances with minimal human effort, 
requiring manual post-processing for fewer than 100 cases.


\subsection{Our Classification-Generation Framework}

We expect our framework to possess the following capabilities:
1) Strong performance on both the IDRR task and the associated explanation generation task, with the two tasks mutually enhancing each other;
2) Explanations that are well-aligned with the model’s classification decisions;
3) Plug-and-play compatibility, enabling seamless integration with most existing IDRR models.
A straightforward strategy is to reformulate the classification task as a generation task, enabling a generation-based model (e.g., T5) to produce a label and its explanation, in flexible order.
However, we identify two main drawbacks of this strategy:
1) it often fails to achieve performance comparable to classification-based or prompt-based IDRR models of similar scale;
and 2) it lacks generality due to challenges in integrating with existing classification models.



\begin{figure}[t]
\centering
\includegraphics[width=7.5cm]{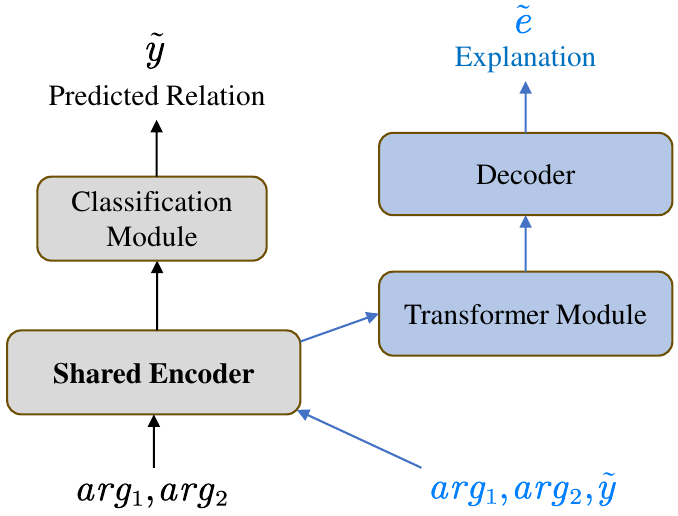}
\caption{Our proposed classification-generation framework. 
For ease of understanding, we use the IDRR task as an example to illustrate our framework.
It first predicts the discourse relation $\tilde y$, which is then used as extra input to generate the explanation $\tilde e$. 
\label{fig:gcmodel}}
\end{figure}

We propose a simple yet effective framework that learns to jointly predict discourse relations and generate explanations, guided by both gold labels and LLM-generated explanations.
As shown in Figure \ref{fig:gcmodel}, 
our framework consists of a shared encoder, a module for classification, 
and a decoder paired with a transformer module for explanation generation.
The two tasks can interact and benefit from each other through the shared encoder.
Following prior work in IDRR, 
we adopt RoBERTa-like PLMs \citep{liu_roberta:_2019} as the shared encoder.
To handle the limited training data,
we employ the decoder component of T5-like PLMs \citep{raffel_exploring_2020} for generation.
However, directly combining the two types of PLMs causes a mismatch, 
as they are pre-trained for different purposes: one for natural language understanding and the other for generation.
We introduce a transformer module—consisting of a randomly initialized transformer layer—to mitigate the mismatch, and find it effective in practice.
To ensure better alignment between classification outputs and generated explanations, we first predict the discourse label and subsequently condition the explanation generation on this label.


Our plug-and-play framework is compatible with both classification-based and prompt-based IDRR models.
We exemplify its integration using prompt-based models due to their superior performance.
Given an instance $(arg_1, arg_2)$,
a prompt-based IDRR model usually calculates the predicted relation $\tilde y$ as follows:
\begin{equation}
\label{eq:classification}
\begin{aligned}
\tilde y &= \text{Verbalizer}(h_{mask}),\\
h_{mask} &= \text{RoBERTa}(T_a(arg_1, arg_2)), 
\end{aligned}
\end{equation}
where $T_a$ is the prompt template for IDRR, 
$h_{mask}$ is the representation of the $\textless mask\textgreater$ token.
RoBERTa is a pre-trained masked language model.
The Verbalizer predicts the most likely discourse connective (e.g., \emph{but}) at the masked position, 
which is subsequently mapped to a discourse relation (e.g., \emph{Comparison}) based on handcrafted rules detailed in \emph{Appendix A}.
In our experiments, we use the template\footnote{Arg1:$arg_1$.Arg2:$arg_2$.$\textless/s\textgreater\textless/s\textgreater$The conjunction between Arg1 and Arg2 is $\textless mask\textgreater$.} defined in \citep{zhou_prompt-based_2022},
and the Verbalizer in \citep{xiang_connprompt_2022}.
We omit the details of Verbalizer for brevity.
During training, we adopt the standard cross-entropy loss for classification (as $L_c$ in Equation \ref{eq:loss}).

For explanation generation, we only need to stack a minimal transformer module and a generation decoder on top of the shared RoBERTa encoder.
Formally, the explanation $\tilde e$ is generated as follows:
\begin{equation}
\label{eq:generation}
\begin{aligned}
\tilde e &= \text{T5-Decoder}(\hat H),\\
\hat H &= \text{Transformers}(H_{last}), \\
H_{last} &= \text{RoBERTa}(T_b(arg_1, arg_2, \tilde y)), 
\end{aligned}
\end{equation}
where $T_b$ is the prompt template\footnote{Arg1:$arg_1$.Arg2:$arg_2$.$\textless/s\textgreater
\textless/s\textgreater$The conjunction between Arg1 and Arg2 is $\tilde y$, the main reason is that.} for explanation generation,
$H_{last}$ is the final layer outputs of RoBERTa,
$\text{Transformers}$ consists of several stacked Transformer layers and $\hat H$ is the adaptive representation for generation.
We adopt the standard auto-regressive cross-entropy loss for text generation as $L_g$ in Equation \ref{eq:loss}.

During training, we adopt a multi-task learning approach.  
Given the prediction $\tilde y$ and the generation explanation $\tilde e$, the total loss $L_{total}$ is defined as: 
\begin{equation}
\label{eq:loss}
\begin{aligned}
L_{total} = \alpha L_{c}(y, \tilde y) + \beta L_{g}(e, \tilde e),
\end{aligned}
\end{equation}
where $L_{c}$ and $L_{g}$ are the respective losses for classification and explanation generation, with $y$ and $e$ as the true label and explanation, 
$\alpha$ and $\beta$ are the coefficients. 
During training, our framework usually exhibits good performance on the classification task after just a few epochs, but struggles with the generation task.
To tackle this issue, 
we first emphasize training on the generation task ($\alpha < \beta$) and then shift focus to the classification task ($\alpha > \beta$).

The key difference from prior frameworks \citep{camburu_e-snli_2018,liu_towards_2019} is that we introduce a Transformer module to bridge the gap between classification and generation modules — a crucial addition when combining separately pre-trained understanding and generation LMs.

\section{Experiments}
\subsection{Dataset and Settings}
To evaluate the effectiveness of our method, we carry out experiments on the PDTB corpus.
Following previous work \citep{wu_label_2022}, 
we divide the corpus into three parts: Sections 2-20 as the training set (12,775 instances), Sections 0-1 as the validation set (1,183 instances), and Sections 21-22 as the test set (1,046 instances).
Four primary top-level discourse relations are considered: \emph{Temporal}, \emph{Comparison}, \emph{Contingency}, and \emph{Expansion}.
More details of these datasets can be found in \emph{Appendix B}.

Our model is trained using a two-stage approach \citep{zhou2022multi}. 
In the first stage, we assign different learning rates: $5e\text{-}6$ for RoBERTa and $5e\text{-}5$ for the other modules.
To emphasize the generation task, we set the loss weights to $\alpha=0.4$ and $\beta=0.6$, and trained for 30 epochs.
In the second stage, we adjust the learning rate for all modules except RoBERTa to $3e\text{-}5$.
The coefficient values $\alpha=0.8$ and $\beta=0.2$ are employed to shift the training emphasis towards the IDRR task.
Our model reaches the optimal validation performance on IDRR within just 6 epochs.
In both stages, we use the AdamW optimizer with a 0.3 dropout rate. 
The Transformer module has one randomly initialized layer, as adding more layers yields no gains. 
Following prior work, we evaluate using macro-averaged $F_1$ score and accuracy ($Acc$), averaging results over three random seeds on an NVIDIA 3090 GPU.

\subsection{Performance of IDRR}

We compare our method (named as EIDRR) with recent baselines, 
which primarily include the following:

\textbf{LLM-based methods:} 
ChatGPT is evaluated under the in-context learning setting \citep{chan_exploring_2024}.
We formulate IDRR as a generation task and fine-tune\footnote{https://github.com/hiyouga/LLaMA-Factory} LLaMA-3B and LLaMA-8B with LoRA. 

\textbf{T5-based methods:} 
We reformulate IDRR as a generation task, utilizing T5 to  produce label text and corresponding explanations sequentially (T5+IDRR$_{gt}$).
We stack a classification layer on the encoder of T5 for IDRR and leverage the T5 decoder to generate explanations (T5+IDRR$_{ct}$) with the predicted labels as additional inputs.


\textbf{Non-prompt-based methods:} 
BMGF \citep{liu_importance_2020}, CVAE \citep{dou_cvae-based_2021}, LDSGM \citep{wu_label_2022}, 
GOLF \citep{jiang_global_2023}, SCIDER \citep{cai-etal-2024-improving}.

\textbf{Prompt-based methods:}
PCP \citep{zhou_prompt-based_2022},
PEMI \citep{zhao_infusing_2023},
and Discoprompt \citep{chan_discoprompt_2023}.


\renewcommand{\arraystretch}{1.15}
\begin{table}[t]
\centering
\begin{tabular*}{\linewidth}{@{\extracolsep{\fill}}c@{}c@{}c@{}}
\toprule
Method & $Acc$ (\%) & $F_1$ (\%) \\
\midrule

ChatGPT & 50.24 & 44.09  \\
LLaMA-3B~+~IDRR$_{gt}$ & 65.11 & 57.11 \\
LLaMA-8B~+~IDRR$_{gt}$ & 70.36 & 61.01 \\

\hline
T5~+~IDRR$_{gt}$~w/o~Exp & 64.05 & 55.92 \\
T5~+~IDRR$_{gt}$ & 64.63 & 56.65 \\

T5~+~IDRR$_{ct}$~w/o~Exp & 63.67 & 52.39 \\
T5~+~IDRR$_{ct}$ & 64.91 & 55.71 \\

\hline
BMGF & 69.06 & 63.39  \\
CVAE & 70.17 & 65.06  \\
LDSGM & 71.18 & 63.73  \\
GOLF & 72.52 & 65.76 \\
SCIDER & 72.11 & 67.00 \\

\hline
PCP & 70.84 & 64.95 \\
PEMI & 71.13 & 64.05 \\
Discoprompt & 71.70 & 65.79 \\

\hline 
EIDRR~w/o Exp & 71.80 &66.87\\
EIDRR & \textbf{73.14} & \textbf{68.01} \\ 
\bottomrule
\end{tabular*}
\caption{Comparison with baselines. 
IDRR$_{gt}$ and IDRR$_{ct}$ means that IDRR is formulated as a generation task and a classification task, respectively.
Here w/ Exp and w/o Exp indicate whether the explanation generation task is included or omitted.}
\label{tab:acc-idrr}
\end{table}

Based on the results in Table \ref{tab:acc-idrr}, 
we can draw the following conclusions. 
\textbf{Firstly}, a powerful model like ChatGPT struggles to achieve promising performance on IDRR without fine-tuning.
While fine-tuned LLMs like LLaMA show some improvements (Part 1), they still lag behind prompt-based IDRR models (Part 4).
\textbf{Secondly}, directly using an LM pre-trained for generation (T5-based methods, Part 2) still fails to obtain comparable performance to current models such as SCIDER and Discroprompt.
The main reason is that LMs pre-trained for generation do not perform as well in classification tasks like IDRR as those pre-trained for language understanding.
\textbf{Thirdly}, including the explanation generation task boosts the IDRR performance across all three settings (w/o Exp). 
These results strongly suggest that explanations generated via LLMs are useful.
\textbf{Lastly}, our EIDRR significantly outperforms the prompt-based Discroprompt, 
which leverages label hierarchy information.
The improvement is primarily attributed to incorporating the explanation generation task in a multi-task learning way, with 1.34\% and 1.14\% gains in $Acc$ and $F_1$, respectively (EIDRR vs. w/o explanation).
A closer look at the results reveals that EIDRR obtains higher $F_1$ scores for each relation.
In addition, results of the ablation studies are provided in \emph{Appendix C}.

Overall, by leveraging LLM-generated explanations as additional supervision, our classification-generation framework effectively improves IDRR performance.


\subsection{Quality Evaluation on Explanation}

\begin{figure}[t]
    \centering
    \begin{subfigure}{\linewidth}
        \centering
        \renewcommand{\arraystretch}{1.3}
        \begin{tabular}{c|c|c|c}
            \hline
            Method & $Acc$ (\%) & $F_1$ (\%) & Human-Avg \\
            \hline
            e-INFERSENT & 73.04 & 65.79 & 1.71  \\
            \hline
            EIDRR & \textbf{73.14} & \textbf{68.01} & \textbf{4.20} \\
            \hline
        \end{tabular}
        \caption{}
    \end{subfigure}
    \begin{subfigure}{\linewidth}
        \centering
        \includegraphics{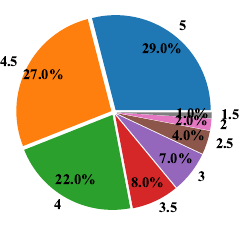}
        \caption{}
    \end{subfigure}
    
    \vspace{0.2cm} 
    \begin{subfigure}{\linewidth}
        \centering
        \includegraphics{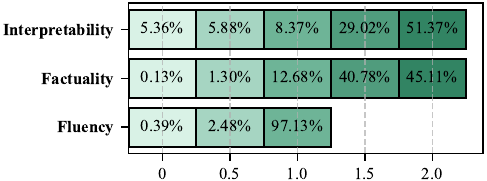}
        \caption{}
    \end{subfigure}
    
    \caption{Quality of generated explanations. 
            (a) Compared with e-INFERSENT, our method achieves a higher average score of 4.20 out of 5.
            (b) The distribution of scores assigned through human evaluation.
            (c) The distribution of scores for each aspect.
            }
    \label{fig:human}
\end{figure}


Following \citet{camburu_e-snli_2018} and \citet{majumder2022knowledge}, 
we restrict the evaluation to correctly predicted test instances, 
since explanations are conditioned on the predicted labels and are unlikely to be correct with wrong labels.
Five human annotators were recruited for the evaluation. 
Each instance was independently scored by two annotators, 
and a third was involved if their scores differed by more than two points. The final score for each instance was computed as the average of the two closest scores.
We asked annotators to evaluate the quality of explanations across three aspects(more details are provided in \emph{Appendix D}):
\begin{itemize}
\item \textbf{Interpretability}: \emph{Does the rationale part justify the predicted label?} We assign scores of 2 (Yes), 1 (Weak-Yes), or 0 (No).
\item \textbf{Factuality}: \emph{Does the restatement part align with the facts stated in the arguments?} We assign scores of 2 (Yes), 1 (Weak-Yes), or 0 (No).
\item \textbf{Fluency}: \emph{Is the explanation grammatically correct and natural?} We assign scores of 1 (Yes) or 0 (No).
\end{itemize}

We compare our framework with e-INFERSENT \citep{camburu_e-snli_2018}, which directly couples the classification encoder with the generation decoder.
It also follows the PredictANDExplain strategy by prepending the predicted label to the input for explanation generation.
For a fair comparison, we substitute the corresponding components in e-INFERSENT with RoBERTa and the T5 decoder, respectively.
From the results in Figure \ref{fig:human} (a), our EIDRR outperforms e-INFERSENT in both IDRR and explanation generation.
The poor explanation quality of e-INFERSENT (average score of 1.71) mainly stems from the incompatibility between the pre-trained T5 decoder and RoBERTa without the transformer module.
A deeper analysis reveals that the explanations generated by e-INFERSENT often fail to align with the input arguments.
In addition, we experimented with the ExplainThenPredict strategy, which first generates explanations and then uses them for classification. 
However, we found that generating explanations without access to labels often leads to inconsistencies, and using such explanations as additional input significantly degrades classification performance.
This experimental observation is consistent with that reported in \citep{camburu_e-snli_2018}.


As shown in Figure \ref{fig:human} (b), 93\% of explanations scored 3 or higher, while only 3\% scored 2 or below. 
More importantly, the average score of 4.20 out of 5 demonstrates that our framework trained with LLM-generated explanations produces fluent and interpretable outputs.
Figure \ref{fig:human} (c) also presents the score distribution across evaluation aspects. In terms of interpretability, 88.76\% of explanations scored 1 or above, indicating that most provide meaningful rationales. 
Regarding factuality, 98.57\% scored 1 or higher, showing that our model effectively captures the key semantics of both arguments. 
The generated explanations are highly fluent, with 97.13\% achieving a perfect score, benefiting from the strong generation capability of the T5 decoder.

Based on the results in Table \ref{tab:acc-idrr} and Figure \ref{fig:human}, 
we conclude that our approach effectively transfers the reasoning capabilities of LLMs to lightweight models, thereby improving IDRR performance while enhancing model interpretability.

\begin{figure}[t]
\centering
\includegraphics[width=8.5cm]{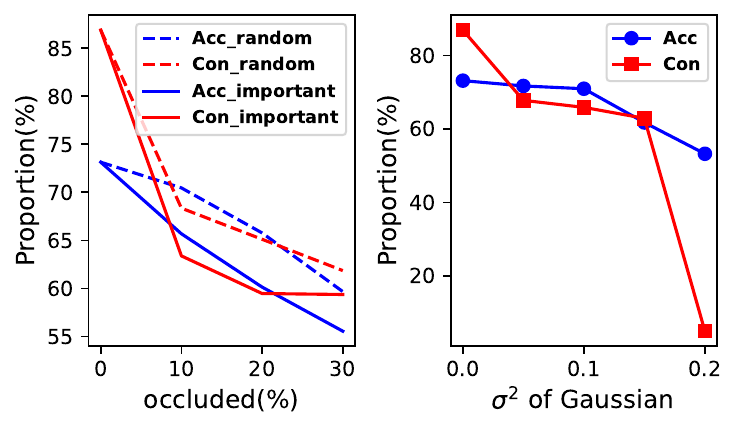}
\caption{Feature Importance Agreement (Left) and Robustness Equivalence (Right). 
\emph{Con} denotes the proportion of label–explanation alignment. 
\emph{Random} and \emph{Important} indicate that randomly selected or top-ranked important features are occluded, respectively.
\label{fig:faithful}}
\end{figure}

\subsection{Faithful Evaluation on Explanation}
It remains unclear whether these explanations genuinely capture the IDRR model’s reasoning process. 
While evaluating the faithfulness of natural language explanations is inherently challenging, we address this by adopting \emph{Feature Important  Agreement} and \emph{Robust Equivalence}  \citep{wiegreffe_measuring_2021,majumder2022knowledge}.


\emph{Feature Importance Agreement}
Features important for classification should also influence explanation generation, and vice versa.
To validate this, we employ a gradient-based attribution method to identify salient input features (tokens) ranked within the top-\{10, 20, 30\}\% attribution scores for the given task. Subsequently, we remove these features and quantify their impact on the model's predictive performance.
As illustrated in Figure \ref{fig:faithful} (Left), the removal of explanation-relevant features (blue solid line) causes a more pronounced reduction in classification accuracy than randomly removing features (blue dashed line). Likewise, eliminating classification-relevant features (red solid line) lowers label–explanation consistency to a greater extent than random removal (red dashed line). 
This indicates mutual importance of features across two tasks and supports the faithfulness of generated explanations.

\emph{Robustness Equivalence}
We inject zero-mean Gaussian noise $\mathcal{N}(0, \sigma^2)$ into internal features and observe changes in classification accuracy and label-explanation consistency. 
The goal is to assess whether both label prediction and explanation generation remain stable (or unstable) under similar noise conditions and exhibit similar trends in performance degradation.
Figure \ref{fig:faithful} (Right) shows that, as the noise level increases, both metrics degrade gradually and in a similar trend, indicating stable model behavior. 
Under high noise (e.g., $\sigma^2=0.2$), both tasks fail: classification accuracy drops to 53.25\%, 
comparable to that of a randomly initialized model, 
and the generated explanations become irrelevant to the input. This shared sensitivity to noise further supports the faithfulness of the generated explanations.

\begin{table*}[h]
    \centering
    \fontsize{9pt}{9pt}\selectfont
    \begin{tabular}{>{\arraybackslash}p{0.41\textwidth}|p{0.54\textwidth}}
        \hline
         \makecell[c]{Example and Predictions} & \makecell[c]{Explanation} \\
        \hline
        \multirow{2}{0.41\textwidth}{\justifying \textbf{Arg1}: Instead, the rally only paused for about 25 minutes and then steamed forward as institutions resumed buying.\\
         \textbf{Arg2}: The market closed minutes after reaching its high for the day. \\ 
         \textbf{Relationship}: Temporal \\
         \textbf{EIDRR}: Temporal\ding{51} 
         \textbf{w/o explanation}: Comparison \ding{55}}
        & \multirow{2}{0.54\textwidth}{\justifying The first sentence describes the slow start of a rally, which then moved forward with institutions resuming buying. The second sentence mentions the closing price of the market, which was just after reaching its highest point for the day. \textbf{The temporal relationship is evident because the second event (the market closing after reaching its highest point) happened after the first event (the rally slowing and then moving forward).} This sequence of events establishes the temporal connection between the two sentences.} \rule{0pt}{53pt}\\
        \hline

        \multirow{2}{0.41\textwidth}{\justifying \textbf{Arg1}: We view the market here as going through a relatively normal cycle \\
         \textbf{Arg2}: We continue to feel that the stock market is still the place to be for long-term appreciation. \\
         \textbf{Relationship}: Expansion \\
         \textbf{EIDRR}: Expansion\ding{51} 
         \textbf{w/o explanation}: Contingency \ding{55}}
        & \multirow{2}{0.54\textwidth}{\justifying The first sentence states that the market is seen as undergoing a relatively normal cycle. The second sentence provides additional information about the market's continued positive outlook for long-term appreciation.  \textbf{The second sentence expands on the idea introduced in the first sentence by giving more detail about the market's outlook and the reasons behind it. } Thus, the relationship between the two sentences is one of expansion.} \rule{0pt}{44pt}\\
        \hline

        \multirow{2}{0.41\textwidth}{\justifying \textbf{Arg1}: It would have been too late to think about on Friday.\\
         \textbf{Arg2}: We had to think about it ahead of time. \\
         \textbf{Relationship}: Contingency \\
         \textbf{EIDRR}: Contingency\ding{51} 
         \textbf{w/o explanation}: Comparison \ding{55}}
        & \multirow{2}{0.54\textwidth}{\justifying The first sentence indicates a situation where thinking about something would have been too late on Friday.  The second sentence states that there was a need to think about it before it could be decided.  \textbf{The contingency relationship is demonstrated here because the need for thinking ahead of time (second sentence) is contingent upon the situation described in the first sentence.}  In other words, the need for thinking ahead is based on the situation being discussed.} \rule{0pt}{52pt}
        \\
        \hline

        \multirow{2}{0.41\textwidth}{\justifying \textbf{Arg1}: Tokyo stocks closed off a significant but less-than-alarming 1.8 \% on thin volume\\
        \textbf{Arg2}: Hong Kong stocks declined 6.5 \% in orderly trading \\
        \textbf{Relationship}: Comparison \\
        \textbf{EIDRR}: Comparison\ding{51} 
        \textbf{w/o explanation}: Expansion \ding{55}}
        & \multirow{2}{0.54\textwidth}{\justifying The first sentence states that Tokyo stocks closed lower by 1.8\% on a narrow volume, which is less than analgamating, but less than analgamating one. The second sentence mentions that Hong Kong stocks declined by 6.5\% in stable trading. \textbf{These two sentences form a comparison by contrasting the closing performance of Tokyo stocks with the performance of HK stocks in different markets.} The comparison highlights the difference in performance between the two stocks in terms of their closing value and their performance in different markets.} \rule{0pt}{60pt}\\
        \hline
    \end{tabular}
    \caption{Examples of IDRR explanations. More examples are in \emph{Appendix E} and on \emph{Github} (all test instances). }
    \label{tab:case-study}
\end{table*}

\begin{table}[t]
    \renewcommand{\arraystretch}{1.25}
    \centering
    \begin{tabular}{llccc}
        \hline
         Task &  & $Acc$ (\%) & $F_1$ (\%)  & Human  \\
        \hline
        \multirow{2}{*}{\centering ATSC} & w/o explanation   & 85.59 & 70.18 & -  \\ 
                        & w/ explanation     & \textbf{87.44 }& \textbf{71.30}  & 4.01 \\
        \hline
        \multirow{2}{*}{\centering NLI}   & w/o explanation  & 74.83 & 74.79 & -  \\ 
                         & w/ explanation    & \textbf{75.83} & \textbf{75.73}  & 4.26 \\
        \hline
    \end{tabular}
    \caption{The performance of ATSC and NLI.}
    \label{tab:atsc-nli}
\end{table}

\subsection{Case Study}

We provide several examples in Table \ref{tab:case-study} to demonstrate the faithfulness of generated explanations, 
For the first example, the generated explanation captures the key \emph{Temporal} cues, 
such as \emph{reaching its highest point happened after the rally slowing and them moving forward}. 
However, the comparison model may have been influenced by the word \emph{instead} in $Arg1$ and incorrectly predicted a comparison relationship.
The second example is very confusing even for humans. Is it \emph{Expansion} or \emph{Contingency}?
EIDRR generated a reasonable explanation as \emph{giving more detail about the market’s outlook and the reasons behind it}.
We can see that the explanation conveys both the \emph{Expansion} relationship through words \emph{more detail} and the \emph{Contingency} relationship through \emph{reasons}.
EIDRR prioritizes predicting as the \emph{Expansion} relationship.
For the third example, the comparison model misclassified it as \emph{Comparison}, possibly due to the presence of words \emph{think about} in both arguments.
EIDRR makes the correct prediction and provides a consistent explanation, stating that \emph{thinking ahead is based on the situation being discussed}.
Similarly, our EIDRR generated a faithful explanation for the \emph{Comparison} relationship in the last instance.
Overall, these examples clearly highlight our method's ability to generate explanations.

\subsection{Performance of ATSC and NLI}

To demonstrate the generality of our method, we evaluate it on aspect term sentiment classification (ATSC) and natural language inference (NLI).
Specifically, we use the Restaurant15 dataset \citep{pontiki-etal-2015-semeval} in our experiment, 
with the training, validation, and test sets containing 963, 361, and 576 instances, respectively.
For the NLI task, we randomly sampled 1,499 training, 400 validation, and 600 test instances from the corpus used in \citep{chen-etal-2017-recurrent-neural}.
The prompt templates used for ATSC and NLI are shown in \emph{Appendix F}.
From the results in Table \ref{tab:atsc-nli}, we can draw the following conclusions. 
\textbf{Firstly}, our method shows improvements of approximately 1\% on two tasks, in terms of both $Acc$ and $F_1$.
Specifically, we achieve 1.85\% gain in $Acc$ and 1.12\% in $F_1$ score on  ATSC.
It is noteworthy that these gains are achieved with only limited explanations.
\textbf{Secondly}, we manually evaluate the generated explanations and obtain average scores exceeding 4.0 on both tasks.  
Examples of ATSC and NLI explanations are shown in \emph{Appendix G}.
Overall, our method performs well on both the ATSC and NLI tasks, 
delivering higher classification performance and high-quality explanations, 
confirming its effectiveness and generality.

\section{Conclusion}

In this study, we successfully transferred the reasoning capabilities of LLMs to enhance both the performance and interpretability of lightweight IDRR models.
Our classification–generation framework can be readily combined with existing classifiers and demonstrates effectiveness not only on the IDRR task but also shows potential for broader applications. 
Future work includes developing more effective methods to evaluate the faithfulness of generated explanations and further exploring the interpretability of model internals in discourse relation recognition \citep{miao_etal_2025_discursive,mondorf_etal_2025_circuit}.

\section{Acknowledgments}
This work was supported in part by the National Natural Science Foundation of China (Nos. 62266017 and
62166018), the Natural Science Foundation of Jiangxi Province of China (Nos. 20232BAB202050 and 20242BAB25117), and Jiangxi
Province Key Laboratory of Advanced Network Computing under Grant No. 2024SSY03071.


\bigskip

\bibliography{aaai2026}

\clearpage

\setcounter{secnumdepth}{0}
\renewcommand\thesubsection{\arabic{subsection}}
\renewcommand\labelenumi{\thesubsection.\arabic{enumi}}

\newcounter{checksubsection}
\newcounter{checkitem}[checksubsection]

\newcommand{\checksubsection}[1]{%
  \refstepcounter{checksubsection}%
  \paragraph{\arabic{checksubsection}. #1}%
  \setcounter{checkitem}{0}%
}

\newcommand{\checkitem}{%
  \refstepcounter{checkitem}%
  \item[\arabic{checksubsection}.\arabic{checkitem}.]%
}
\newcommand{\question}[2]{\normalcolor\checkitem #1 #2 \color{blue}}
\newcommand{\ifyespoints}[1]{\makebox[0pt][l]{\hspace{-15pt}\normalcolor #1}}

\end{document}